# Saliency Map Estimation for Omni-Directional Image Considering Prior Distributions

Tatsuya Suzuki
Department of Information and Communication Sciences
Sophia University
Tokyo, Japan
t-suzuki-7mf@eagle.sophia.ac.jp

Takao Yamanaka
Department of Information and Communication Sciences
Sophia University
Tokyo, Japan
takao-y@sophia.ac.jp

*Abstract*—In recent years, the deep learning techniques have been applied to the estimation of saliency maps, which represent probability density functions of fixations when people look at the images. Although the methods of saliency-map estimation have been actively studied for 2-dimensional planer images, the methods for omni-directional images to be utilized in virtual environments had not been studied, until a competition of saliency-map estimation for the omni-directional images was held in ICME2017. In this paper, novel methods for estimating saliency maps for the omni-directional images are proposed considering the properties of prior distributions for fixations in the planar images and the omni-directional images.

*Keywords—omni-directional image, saliency map, deep learning*

## I. INTRODUCTION

Omni-directional cameras are expected to be utilized in widespread applications such as virtual environments, robotics, and surveillance systems. The estimation of the saliency map for an omni-directional image (ODI), that is the probability distribution of the gazing points when people look at ODI with a head-mounted display (HMD), will be useful for a variety of applications including ODI image compression, setting of the initial viewing direction for displaying ODI, creating a movie for summarizing ODI, and detection of interesting objects in virtual/augmented environments. Examples of the saliency maps for 2-dimensional (2D) planar images and ODI are shown in Figure 1.

While the deep learning techniques have greatly improved the accuracy of saliency-map estimation for 2D images [1, 2, 3], there are few studies to estimate the saliency maps for ODI. The competition of saliency-map estimation for ODI held in ICME2017 [4, 5, 6, 7] is the first attempt for this topic. It has been known that people tend to gaze at the center in a 2D image (center bias) [8] and the equator in ODI (equator bias) [9]. Therefore, the prior distribution needed to estimate the saliency maps for ODI is different from that for the 2D image, though these differences of prior distributions has not been clearly considered in the previous studies [7, 10]. In this paper, novel methods for estimating saliency maps for ODI are proposed considering the difference of the properties for prior distributions in 2D images and ODI.

The contributions of this paper include:
1) Developing saliency-map estimation models for ODI considering the difference of prior distributions for 2D images and ODI.
2) Learning equator bias for ODI, explicitly.
3) Improving accuracy of saliency-map estimation for ODI by a large margin.

## II. RELATED WORK

In the last 20 years, many saliency models have been proposed to estimate the locations in 2D images which attract attentions. Although most of the conventional models have used low-level features extracted by edge-detectors, color filters, and local image statistics [11, 12], the convolutional neural network models based on deep learning have greatly improved the accuracy of the saliency-map estimation. For example, SaliconNet [1] and DeepGazeII [2] based on the VGG architecture [13] are the state-of-the-art models for estimating saliency maps of 2D images in the MIT Saliency Benchmark [14]. Furthermore, DenseSal [3] showed better performance based on densely connected neural networks [15] instead of the VGG architecture.

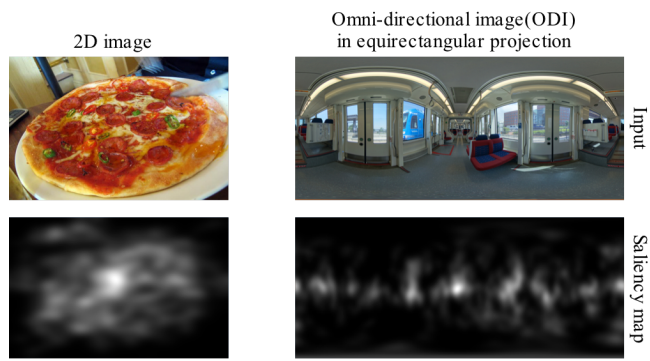

Figure 1: Examples of saliency maps for 2D image and omni-directional image.

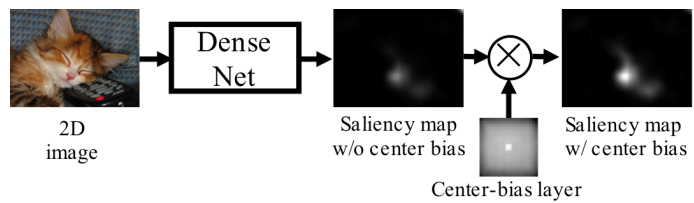

Figure 2: Model for estimating saliency maps for 2D images (DenseSal).

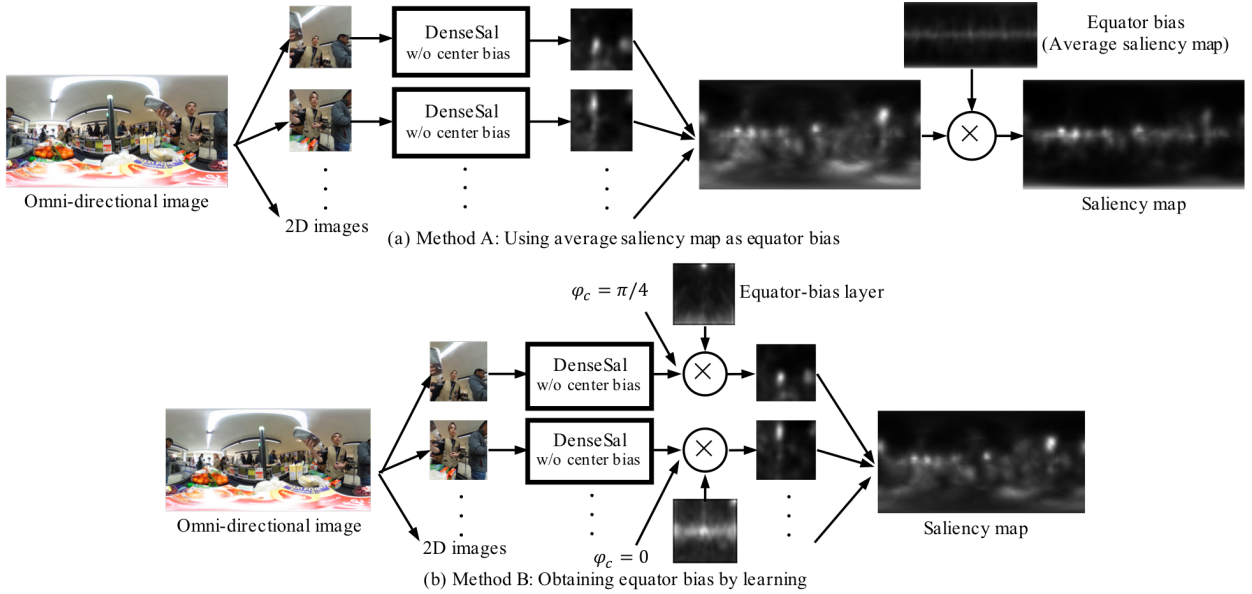

Figure 3: Proposed methods for estimating saliency maps for omni-directional images.

In addition to the saliency-map estimation for 2D images, that for ODI have just started since the ICME2017 competition [4, 5]. In SalNet360 [7], the saliency maps for ODI are estimated using the model based on SaliconNet (2D saliency model) for extracted 2D images from ODI and the refinement neural network to integrate the estimated 2D saliency maps into an equirectangular projected image. In order to incorporate the dependence of saliency on the locations in ODI, the ODI coordinates in the 2D saliency map were input to the refinement neural network. However, since SaliconNet outputs a saliency map biased in the center of an image, the saliency at the end of extracted 2D images is estimated lower. Another saliency-map estimation model for ODI has been proposed by Abreu et al. [10], where SaliconNet is used to estimate the ODI saliency map by inputting directly an equirectangular image projected from ODI. The equirectangular saliency maps for horizontally different viewing directions are fused into an equirectangular saliency map to suppress the center-bias effect induced by the 2D saliency-map model (SaliconNet). However, the equirectangular projected images have distortions at poles, so that saliency at the poles cannot be correctly estimated. Moreover, the center-bias effect cannot be completely suppressed at both ends of the equirectangular projected image. As another saliency-map estimation model for ODI, SaltiNet has also been proposed by Assens et al. [16]. The model is based on a temporal-aware novel representation of saliency information named the saliency volume, which is composed of feature maps with the temporal axis. However, this model does not consider the dependence of saliency on the locations in ODI.

In this paper, novel saliency-map estimation models for ODI are proposed by using a 2D saliency model, by considering the difference of prior distributions for 2D images and ODI, so that this model can estimate the saliency depending on the locations in ODI.

III. METHOD

In the proposed models, the 2D saliency-map estimation model, DenseSal [3], is used with an additional layer for learning the center bias in 2D saliency maps, as shown in Figure 2. This center-bias layer consists of weights to be multiplied by pixel values of the feature map from DenseNet in Figure 2. By learning the weights from training data, the saliency depending on the location in the image can be estimated.

The proposed models for the ODI saliency-map estimation are shown in Figure 3. First, the 2D images are extracted from ODI, feeding into the 2D saliency-map estimation model (DenseSal) in Figure 2. The 2D saliency maps before the center-bias layer from DenseSal (saliency map w/o center bias) for multiple viewing directions are combined to synthesize an equirectangular saliency map for ODI, as shown in Figure 3. In this paper, two ODI saliency-map estimation models are proposed, named Method A and Method B, which are different in how to incorporate the equator bias in the model. Since the prior distribution of fixations for ODI (equator bias) is different from that for 2D images (center bias), the 2D saliency map before center-bias layer (saliency map w/o center bias) is used in both the methods A and B, to exclude the center-bias effect for a 2D image. In Method A, the equator bias is obtained by averaging the saliency maps for training data, to be multiplied by the equirectangular saliency map after the integration of 2D saliency maps for multiple viewing directions, as shown in Figure 3(a). In Method B, the equator bias for each vertical viewing direction is learned from the training data before the integration, in the equator-bias layer in Figure 3(b).

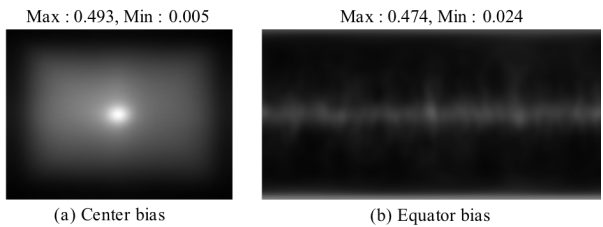

Figure 4: Prior distributions for 2D images and omni-directional images.

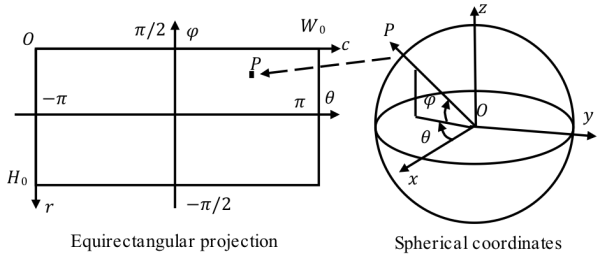

Figure 5: Coordinate systems of omni-directional image.

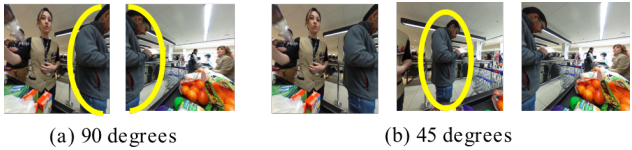

(a) 90 degrees  (b) 45 degrees

Figure 6: Example images of different intervals of viewing directions. A human, who is cut off at the end of the image in the viewing-direction interval of 90 degrees (a), can be detected in the interval of 45 degrees (b).

*A. Center bias and equator bias (prior distributions)*

As shown in the examples of Figure 1, the center bias represents the property of tendency for fixations to concentrate on the center of a 2D image, while the equator bias represents that to concentrate on the equator of ODI. The averages of saliency maps for 2D images and ODI are shown in Figure 4 (a) and (b), respectively. They represent the prior distributions of fixations for those images, independent on the local image features. Thus, the prior distributions are different between the 2D images and ODI, so that these differences have to be considered in the ODI saliency-map estimation.

The saliency is defined as the probability of observing a fixation (*fix*: binary random variable) at a location *x* for a local image feature *f*: $p(fix|x, f)$. If *x* and *f* are assumed to be independent, $p(fix|x, f)$ is proportional to the value of the saliency map without the prior distribution $p(fix|f)$ depending only on the local image feature *f*, and the prior probability $p(x|fix)$: $p(fix|x, f) \propto p(fix|f)p(x|fix)$. Since the DenseSal in Figure 2 is trained with the saliency maps obtained by blurring the observed fixations, the saliency maps output from the model include the center-bias effect (saliency map w/ center bias). Therefore, the saliency maps before the center-bias layer (saliency map w/o center bias) are used for 2D images extracted from ODI. The saliency maps without center bias are multiplied by the equator bias after and before the integration in Methods A and B, respectively. Although the equator bias is multiplied in different steps in Methods A and B, the fundamental difference between the methods is how to obtain the equator bias: averaging the saliency maps for training data in Method A, and learning the equator bias with the equator-bias layer using training data in Method B.

*B. Extraction of 2D images from omni-directional image*

Since the equirectangular image for ODI has the distortion at poles as shown in Figure 1, the 2D planar images are extracted from ODI to estimate the ODI saliency map without the distortion. The correspondence between the equirectangular projection and the spherical coordinate system is shown in Figure 5. The unit vectors of the 2D planar image are represented in the following equations in the 3D Euclidean coordinate system:

$$\begin{aligned} X_n &= (-\sin\theta_c, -\cos\theta_c, 0) \\ Y_n &= (-\sin\varphi_c \cos\theta_c, \sin\varphi_c \sin\theta_c, \cos\varphi_c) \end{aligned} \quad (1)$$

where $(\theta_c, \varphi_c)$ represents the viewing direction for the extracted 2D planar image. The coordinates of the 2D planar image in the equirectangular projection can be obtained by transforming the 3D Euclidean coordinates of the points in the 2D planar image to spherical coordinates. Thus, the 2D planar images can be extracted from the equirectangular image of ODI using the coordinates. When the 2D planar images are integrated into the equirectangular image, each point of the equirectangular image is assigned to the nearest point in 2D planar images of multiple viewing directions.

When the 2D planar images are extracted from ODI without overwrapping (for example, surfaces of cube inscribed in a sphere), the objects placed at the end of the planar image are cut off as shown in Figure 6(a), so that it would be difficult to be recognized for accurate saliency estimation. However, if the 2D images are extracted from ODI with overwrapping as shown in Figure 6(b), the objects can be detected, leading to more accurate estimation. Thus, the proposed models extract 2D images with overwrapping. In the integration of 2D planar saliency maps for multiple viewing directions, the saliency values in equirectangular projection are calculated by averaging the overwrapping saliency values.

*C. Normalization of saliency map*

Since the saliency map is a probability distribution, the 2D saliency-map estimation model such as the model in Figure 2, outputs the saliency map with the normalization so that the sum equals 1. However, when the saliency maps for 2D planar images extracted from ODI are estimated, the probability of fixations depends on the viewing direction for the extraction, so that the information of the dependence is lost by the normalization of each saliency map from the 2D estimation model. Therefore, each 2D saliency map for the 2D image extracted from ODI is not normalized in the model of Figure 2. After the integration of the saliency maps for 2D images into an equirectangular image, it is normalized so that the sum equals 1.

*D. Gaussian filter*

A standard practice for evaluation of 2D saliency maps is to blur the saliency maps with Gaussian filters, and find the optimal size of the Gaussian filter for each model [1]. Therefore, in the evaluation of this paper, the saliency maps for ODI is also blurred by the Gaussian filter with the optimal size based on the evaluation metric defined in Eq. 2 (IV. Experimental Setup)

IV. EXPERIMENTAL SETUP

In the experiments, 'head+eye based saliency maps' in the Salient360! Database [4, 5, 6] created for the ICME2017 competition were used to evaluate the saliency maps for ODI. This database is composed of 40 images for training and 25 images for test in the equirectangular projection with fixations for 17 observers obtained using HMD, Oculus-DK2, whose field of view is 100 degrees.

Table 1: Means and standard deviations for calculating integrated metric $a$ in Eq. 2.

|       | KL    | CC    | NSS   | AUC   |
|-------|-------|-------|-------|-------|
| $m_j$ | 0.400 | 0.623 | 0.806 | 0.713 |
| $\sigma_j$ | 0.035 | 0.055 | 0.072 | 0.016 |

The 2D saliency-map model used in this paper shown in Figure 2 was first trained with the ImageNet classification task, followed by the 2D saliency-map estimation task with Salicon Dataset and OSIE Dataset, which was the same procedure as that in the reference [1, 3]. This model was further fine-tuned with the extracted 2D images for multiple viewing directions from ODI of training data in Salient360 Dataset, where the 32 images in equirectangular projection were used for training, while the remaining 8 images were used for validation. For the extraction of 2D images from ODI, the angle of view was set to 100 degrees, same as the field of view in HMD. The vertical and horizontal viewing directions for the 2D images were set in the constant intervals depending on the number of extracted 2D images from ODI. When the number of extracted 2D images was 6 images (4 viewing directions on equator and 2 directions at poles) for each ODI, the interval of the viewing directions was set to 90 degrees. Similarly, the intervals were set to 45, 30, and 22.5 degrees for 26, 62, and 114 images, respectively. The size of ODI in equirectangular projection was 800×1600 pixels, whereas the extracted 2D images were 500×500 pixels.

The saliency maps of ODI were evaluated with 4 different types of metrics, Kullback-Leibler divergence (KL), Pearson's correlation coefficient (CC), normalized scanpath saliency (NSS), and area under receiver operating characteristic curve (AUC), following the reference [7]. KL and CC are the metrics to measure the difference of probabilistic distributions by comparing the estimated saliency maps with the ideal saliency maps created by blurring the observed fixations with a Gaussian filter. NSS and AUC are the metrics based on the saliency values at the observed fixations. Higher values represent better performance of the saliency-map estimation for CC, NSS, and AUC, whereas lower values represent better performance for KL.

Since these 4 types of metrics measure different characteristics of the saliency-map estimation, an integrated metric was used in the experiments for the objective comparison by averaging the standardized metrics:

$$a = \frac{1}{4}\left(-\frac{KL - m_{kl}}{\sigma_{kl}} + \frac{CC - m_{cc}}{\sigma_{cc}} + \frac{NSS - m_{nss}}{\sigma_{nss}} + \frac{AUC - m_{auc}}{\sigma_{auc}}\right) \quad (2)$$

where $m_j$ and $\sigma_j$ are the mean and the standard deviation for each metric ($j$=kl, cc, nss, auc) in the proposed models. These values used in the experiments are shown in Table 1. In this integrated metric a, higher value represents better performance.

The Gaussian filters for blurring the saliency maps were applied with the size of 8, 16, 24, 32, 40, and 48 pixels in the standard deviations. The optimal saliency maps were selected for each evaluated method from the maps blurred by the filters above in addition to that without blurring, based on the objective metric defined in Eq. 2.

## V. RESULTS

### A. Comparison of Gaussian-filter sizes

The performance of ODI saliency-map estimation was compared against the different levels of blurring using the Gaussian filters. The results are shown in Figure 7 for the proposed methods A and B with the metrics: (a) KL (inverted), (b) CC, (c) NSS, (d) AUC, and (e) the integrated metric, a. The interval of the viewing directions for the 2D image extraction from ODI was set to 45 degrees (26 images for each ODI). It can be seen from the figures that the estimation accuracy was better as the size of Gaussian filters became larger for the distribution-based metrics (KL and CC), whereas the accuracy was better as the size became smaller for the location-based metrics (NSS and AUC). In the integrated metric a, the accuracy was the highest at the 24 pixels and 0 pixels (without blurring) for Methods A and B, respectively. In the following experiments, the optimal filter size was applied to the estimated saliency maps for each method.

### B. Comparison of viwing-direction intervals

The comparison of the performance against the interval of viewing directions for the 2D image extraction from ODI is shown in Figure 8. As the interval of the viewing directions decreases, the number of 2D images becomes larger with increasing overwrapping area of 2D images since the angle of view for the 2D images is constant. It is seen from the figure that the performance with the viewing-direction interval smaller than 45 degrees outperformed the performance with 90 degrees (6 extracted images). This would be because the objects which are cut off at the end of the extracted 2D image without overwrapping can be detected with overwrapping, as explained in Figure 6. Since the performance was not improved with smaller interval than 45 degrees, the interval of 45 degrees with 26 extracted images were used in the following experiments.

### C. Comparison of methods with prior distributions

The proposed methods (Methods A and B) were compared with the models in the following conditions:

1) The 2D saliency-map model, DenseSal[4], with center-bias layer (w/ CB), without fine-tuning to ODI, for estimating saliency maps of 2D extracted images from ODI.
2) The model same as (1), except for using feature maps before the center-bias layer (w/o CB) and multiplying equator bias obtained by averaging the saliency maps for traning data (w/ EB(average)). Method A without fine-tuning.
3) The model same as (1), but with fine-tuning to ODI. The model similar to Method B but learning a common equator bias for different vertical viewing directions.
4) The model same as (1), except for using feature maps before the center-bias layer (w/o CB), with fine-tuning to ODI.

In addition, the conventional method, SalNet360 [7], and the average of all saliency maps for training data (equator bias) were compared as baselines. The results are shown in Table 2.

As can be seen from the table, Methods A and B outperformed the conventional method, SalNet360[7], for all the metrics. Compared with Method B, the accuracy of Method A was higher in the distribution-based metrics (KL and CC), but lower in the location-based metrics (NSS and AUC). In the integrated metric a, Method A was the best among all the compared methods including the baselines, followed by Method B. Thus, it was found that the saliency maps for ODI can be accurately estimated using the proposed methods. Although the performance in the 2D saliency map model without fine-tuning (1) was lower than the conventional method [7], the model (2) excluding center bias and including equator bias had the performance comparable with the conventional method even without fine-tuning. Furthermore, the performance was greatly improved in the models (3) and (4) because of the fine-tuning to the ODI dataset.

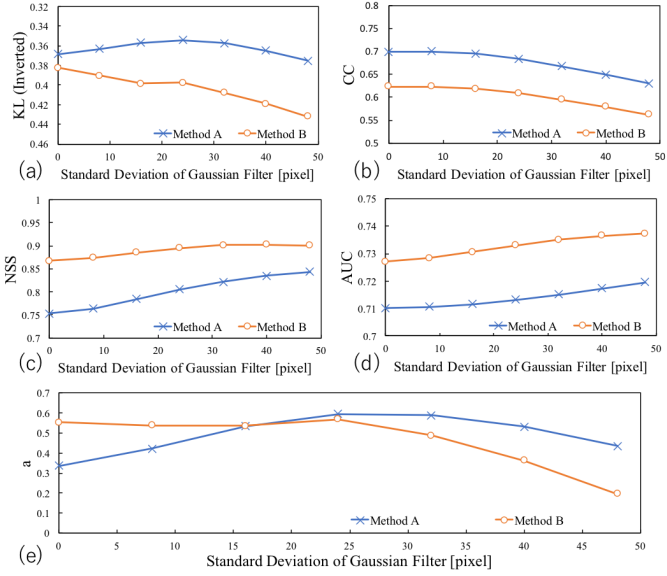

Figure 7: Comparison of saliency estimation accuracy against Gaussian filter size.

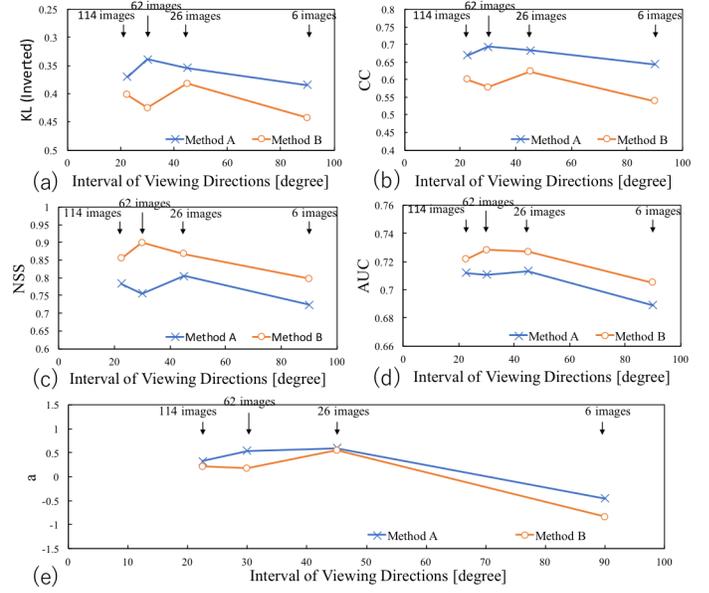

Figure 8: Comparison of saliency estimation accuracy against number of extracted images from omni-directional image.

Examples of the estimated saliency maps are shown in Figure 9, where (a) and (b) show the accurately and inaccurately estimated examples, respectively. In the accurately estimated examples, the model was able to predict the area of high saliency. In the inaccurately estimated examples, it was difficult to estimate the saliency in the area of the building with high saliency, in addition to the chair which was not focused on by observers. In Method A, it was difficult to estimate the saliency in the areas which attract attention at the top and bottom regions of the equirectangular images, because the equator bias obtained by averaging saliency maps of training data had low saliency values around those regions (min = 0.024) as shown in Figure 4(b). On the other hand, the saliency in those regions can be estimated in some degree by Method B.

### D. Equator bias obtained by learning

In Method B, the equator bias for each vertical viewing direction was learned from the training data. In the case of 26 extracted images (interval of 45 degrees), the equator biases for the 5 different vertical directions were obtained, as shown in Figure 10(a). The equator biases at $\varphi_c = -\pi/2$ and $\pi/2$ represent the biases for the 'south' pole and the 'north' pole of viewing directions, respectively, where the fixations were concentrated on the center of those poles. On the other hand, the equator bias at $\varphi_c = 0$ represents the bias for the equator direction, where the fixations were concentrated on the equator corresponding to the horizontal line.

The equator bias integrated in the equirectangular projection is shown in Figure 10(b). The bias learned from the training data was similar to that obtained by averaging the saliency maps in the training data shown in Figure 4(b). However, the range of saliency values in Figure 4(b) (0.024-0.474) was largely different from that in Figure 10(b) (0.915-0.958), which almost equals 1 at all the pixels. This means that the influence of the equator bias in Method B was less than in Method A. This may

Table 2: Evaluation of proposed methods.
FT: Fine-Tuning, CB: Center Bias, EB: Equator Bias, ↓ and ↑ : Directions of higher accuracy

|  |  |  |  | KL ↓ | CC ↑ | NSS ↑ | AUC ↑ | a ↑ |
|---|---|---|---|---|---|---|---|---|
| Baseline | Equator bias (average saliency map) |  |  | 0.441 | 0.588 | 0.366 | 0.639 | -3.124 |
|  | SalNet360[7] | w/ FT |  | 0.458 | 0.548 | 0.755 | 0.701 | -1.116 |
| Compared methods | (1) DenseSal[3] | w/o FT | w/ CB, w/o EB | 1.960 | 0.456 | 0.711 | 0.704 | -12.23 |
|  | (2) |  | w/o CB, w/ EB (average) | 0.672 | 0.581 | 0.795 | 0.724 | -1.976 |
|  | (3) | w/ FT | w/ CB[*1] | 0.383 | 0.613 | 0.852 | 0.724 | 0.525 |
|  | (4) |  | w/o CB, EB | 0.399 | 0.602 | **0.890** | **0.729** | 0.446 |
| Proposed methods | Method A | w/ FT | w/o CB w/ EB (average) | **0.354** | **0.683** | 0.805 | 0.713 | **0.594** |
|  | Method B |  | w/o CB w/ EB (learned) | 0.382 | 0.623 | 0.867 | 0.727 | 0.553 |

*1 Learning a common equator bias for vertical directions using the CB layer.

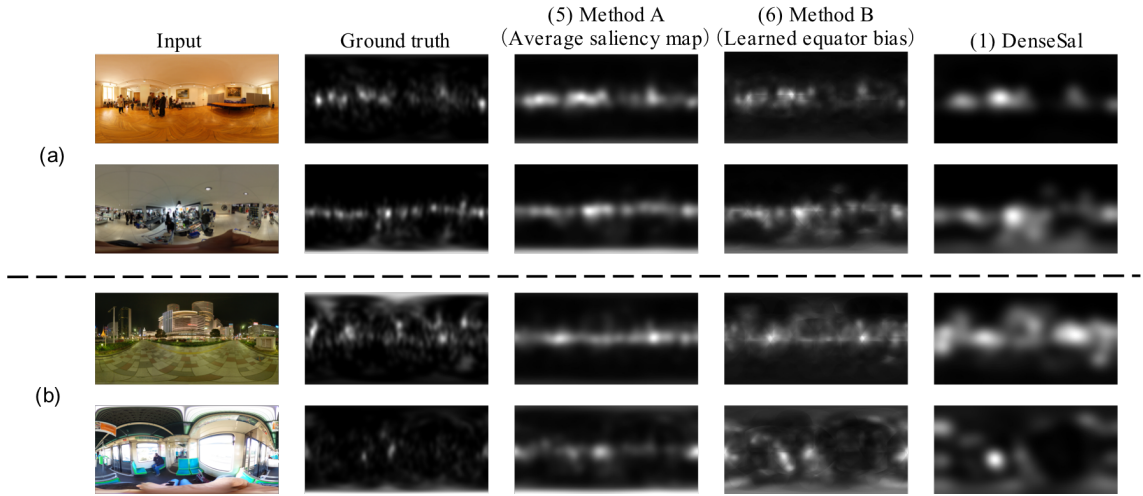

Figure 9: Examples with (a) high accuracy and (b) low accuracy in saliency-map estimation.

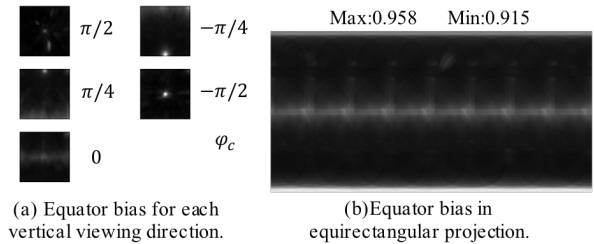

(a) Equator bias for each vertical viewing direction.

(b) Equator bias in equirectangular projection.

Figure 10: Equator bias learned in Method B.

be one of the reasons why Method A was better than Method B in the evaluation, and why even the model (4) without any bias was able to achieve relatively high performance. Although there is no large-scale ODI database sufficient to learning neural networks for ODI saliency-map models, the equator bias will be more accurately learned if the large-scale database is available.

## VI. CONCLUSIONS

In this paper, novel methods for estimating saliency maps for omni-directional images were proposed. By considering the difference between prior distributions of fixations in 2D images and omni-directional images, the proposed methods were able to improve the accuracy of the saliency-map estimation from the conventional method. Moreover, it was found that the extraction of 2D images from an omni-directional image with overwrapping was important for the accurate saliency-map estimation. From the experimental results in this paper, the accuracy of Method A with the equator bias obtained by averaging saliency maps was higher than that of Method B with the bias learned from traning data. However, in the qualitative comparison, Method B output better saliency maps in some cases because the equator bias of the average saliency map in Method A greatly suppressed the saliency values at the top and bottom regions. It will be possible to learn more accurate equator bias in Method B in the future if the sufficient training data is available for the saliency-map estimation in omni-directional images. Furthermore, the metrics for evaluating saliency maps of omni-directional images need to be studied though the metirccs used in the current studies were those for 2D saliency map estimation. Since the top and bottom regions of an equirectangular image are enlarged, the metrics for 2D saliency map estimation cannot correctly be applied for those regions.